\documentclass[conference]{IEEEtran}
\IEEEoverridecommandlockouts
\usepackage{cite}
\usepackage{amsmath,amssymb,amsfonts}
\usepackage{algorithmic}
\usepackage{graphicx}
\usepackage{textcomp}
\usepackage{xcolor}
\usepackage{url}
\usepackage{algorithm}
\usepackage{algorithmic}
\usepackage{graphicx}
\usepackage{amsmath}
\usepackage{graphicx}  
\usepackage{subfig}    
\usepackage{caption}  
\usepackage{color}
\usepackage{framed}
\usepackage{comment}
\usepackage{tcolorbox}
\tcbuselibrary{skins,breakable}
\newtheorem{Theorem}{Theorem}[section]
\newtheorem{Lemma}[Theorem]{Lemma}

\def\BibTeX{{\rm B\kern-.05em{\sc i\kern-.025em b}\kern-.08em
    T\kern-.1667em\lower.7ex\hbox{E}\kern-.125emX}}
\begin{document}

\title{Federated Structured Sparse PCA for Anomaly Detection in IoT Networks\thanks{This work was supported by the National Natural Science Foundation of China under Grant 12371306. (\textit{Corresponding author: Xianchao Xiu.})} 
}

\author{\IEEEauthorblockN{Chenyi Huang}
\IEEEauthorblockA{\textit{School of Mechatronic Engineering and Automation} \\
\textit{Shanghai University}\\
Shanghai, China \\
huangchenyi@shu.edu.cn}
\and
\IEEEauthorblockN{Xianchao Xiu}
\IEEEauthorblockA{\textit{School of Mechatronic Engineering and Automation} \\
\textit{Shanghai University}\\
Shanghai, China \\
xcxiu@shu.edu.cn}
}

\maketitle

\begin{abstract}
Although federated learning has gained prominence as a privacy-preserving framework tailored for distributed Internet of Things (IoT) environments, current federated principal component analysis (PCA) methods lack integration of sparsity, a critical feature for robust anomaly detection.  To address this limitation, we propose a novel federated structured sparse PCA (FedSSP) approach for anomaly detection in IoT networks. The proposed model uniquely integrates double sparsity regularization: (1) row-wise sparsity governed by $\ell_{2,p}$-norm with $p\in [0,1)$ to eliminate redundant feature dimensions, and (2) element-wise sparsity via  $\ell_{q}$-norm with $q\in [0,1)$ to suppress noise-sensitive components. To solve this nonconvex problem in a distributed setting, we devise an efficient optimization algorithm based on the proximal alternating minimization (PAM). Numerical experiments validate that incorporating structured sparsity enhances both model interpretability and detection accuracy. Our code is available at \url{https://github.com/xianchaoxiu/FedSSP}.
\end{abstract}


\section{Introduction}

Internet of Things (IoT) provides us with great intelligence and has become an indispensable part of our lives \cite{le2024applications}. However, the rapid development of IoT networks has also brought new challenges, such as security and privacy \cite{kong2022edge}. To ensure the security of IoT networks, anomaly detection has been widely studied with the aim of identifying potential security threats, see \cite{trilles2024anomaly,inuwa2024comparative,adhikari2024recent} for recent surveys.

Principal component analysis (PCA), as an outstanding dimensionality reduction technique, has been broadly considered in anomaly detection due to its unsupervised nature and effectiveness \cite{xiu2020laplacian,gewers2021principal}. The key idea behind is that anomalies are different from normal behaviors, i.e., they are likely to have larger reconstruction errors compared to normal  observations  \cite{qi2021fast,greenacre2022principal}. In many modern applications, especially in distributed systems like IoT networks, data is spread across multiple local gateways \cite{froelicher2023scalable}. In such settings, traditional PCA is often impractical. To address these challenges, a distributed PCA model is employed, which can be written as
\begin{equation}
       \begin{aligned}
               \min_{W}~~ & \sum_{t=1}^N \| (I - WW^{\top})X_t \|_\textrm{F}^2  \\
               \rm{s.t.}~~ &W^{\top} W=I,
       \end{aligned}
\end{equation}
where $ X_t\in\mathbb{R}^{d\times n} $ is the data at the $t$-th local gateway, and $ W\in\mathbb{R}^{d\times m} $ is the global variable that needs to be learned, $I\in\mathbb{R}^{d\times d} $ denotes the identity matrix. The objective function measures the reconstruction error of the data across all local gateways. However, this model suffers from data redundancy and requires all data to be aggregated to a central server, which is not feasible in many practical applications \cite{qu2024distributed}. 
To this end, Nguyen et al. \cite{nguyen2024federated} proposed a federated PCA model, called FedPG, which is formulated as
\begin{equation}
   \begin{aligned}
           \min_{W_t, Z }~~ & \sum_{t=1}^N \| (I - W_tW_{t}^{\top})X_t \|_\textrm{F}^2  \\
           \rm{s.t.}~~ &W_t = Z, \quad \forall t \in [N]\\
           &W_t^{\top} W_{t}=I, \quad \forall t \in [N],
   \end{aligned}
\end{equation}
where  $[N]=\{1,2,\ldots, N\}$, $ W_t $ is the local variable learned at the $ t $-th local gateway, and $ Z $ is the global variable that needs to be learned. By introducing the global variable $ Z $, FedPG allows local gateways to learn the global variable without sharing their data.
However, it does not consider the sparsity of data, which is crucial for anomaly detection as verified in \cite{xiu2022sparsity,xiu2022efficient,liu2025star}. \textit{A natural question is whether we can integrate sparsity into the federated PCA framework.}

In this paper, we propose the following federated structured sparse PCA (FedSSP) model
\begin{equation}\label{bi-sparse}
    \begin{aligned}
            \min_{W}~~ & \sum_{t=1}^N \| (I - WW^{\top})X_t \|_\textrm{F}^2 +\lambda_1 \|W \|_{2,p}^{p} + \lambda_2 \|W \|_{q}^{q}   \\
            \rm{s.t.}~~ &W^{\top} W=I,
    \end{aligned}
\end{equation}
of which $\lambda_1 $ and $ \lambda_2 $ are regularization parameters. $\|W \|_{2,p}^{p}$ is the $\ell_{2,p}$-norm, defined as $\sum_{i=1}^d \| \mathbf{w}^i\|_2^p$, indicating row-wise sparsity, where $\mathbf{w}^i$ represents the $i$-th row \cite{liu2024efficient,dong2025unsupervised}. $\|W \|_{q}^{q}$ is the $\ell_q$-norm, defined as $\sum_{i=1}^d \sum_{j=1}^m|w_{ij}|^q$, characterizing element-wise sparsity \cite{xiu2018iterative}. More specifically, $p$ and $q$ are within the range of $[0,1)$, which is a unified formulation that seamlessly integrates $\ell_{0}$-norm  and $\ell_{p}$-norm  with $0<p<1$ \cite{zhang2022structured}. The objective function measures the data reconstruction error across all local gateways, while also considering the sparsity of data. 
Overall, the contributions of this paper can be summarized as follows.

\begin{itemize}
\item We propose a novel federated learning framework, which introduces structured sparsity to consider local gateways.
    \item We develop an efficient optimization algorithm using the conjugate gradient method on the Grassmann manifold.
   \item We conduct extensive experiments on real-world datasets to evaluate the performance of the proposed framework.
\end{itemize}

\section{Optimization Algorithm}

In order to solve this problem in a consensus framework, it is necessary to introduce a global variable $ Z $ that is shared among all local gateways. In this regard, our proposed FedSSP can be reformulated as 
\begin{equation}
   \begin{aligned}
           \min_{ W_t,Z }~~ & \sum_{t=1}^N \{ \| (I - W_tW_t^{\top})X_t \|_\textrm{F}^2 \\
          & +\lambda_1 \|W_t\|_{2,p}^{p} + \lambda_2 \|W_t\|_{q}^{q} \}  \\
           \rm{s.t.} ~~ & W_t^{\top} W_t=I, \quad \forall t \in [N]\\
           &W_t = Z, \quad \forall t \in [N].
   \end{aligned}
\end{equation}

Furthermore, we introduce two auxiliary variables $ U_t$ and $V_t$, and rewrite the above problem as
\begin{equation}
    \begin{aligned}
            \min_{ W_t, U_t, V_t, Z }~~ & \sum_{t=1}^N \{\| (I - W_tW_t^{\top})X_t \|_\textrm{F}^2 \\
            &+ \lambda_1 \|V_t\|_{2,p}^{p}+\lambda_2 \|U_t\|_{q}^{q} \} \\
            \rm{s.t.}~~~~~~ &W_t^{\top} W_t=I, \quad \forall t \in [N]\\  
            &V_t = W_t, \quad \forall t \in [N] \\
                  &U_t = W_t, \quad \forall t \in [N]\\
                   &W_t = Z, \quad \forall t \in [N],
            \end{aligned}
\end{equation}
which can be transformed into an unconstrained optimization problem as follows
\begin{equation}\label{bi-sparse3}
	\begin{aligned}
		\min_{ W_t, U_t, V_t, Z }~ & \sum_{t=1}^N \{\| (I - W_tW_t^{\top})X_t \|_\textrm{F}^2 + \lambda_1 \|V_t\|_{2,p}^{p} \\
		&+ \lambda_2 \|U_t\|_{q}^{q} + \Phi(W_t)+ \frac{\beta_1}{2} \| W_t - U_t \|_\textrm{F}^2  \\
		&+ \frac{\beta_2}{2} \| W_t - V_t \|_\textrm{F}^2+ \frac{\beta_3}{2} \| W_t - Z \|_\textrm{F}^2\} ,
	\end{aligned}
\end{equation}
where $\Phi(W_t)$ is the  indicator function defined as
\begin{equation}
	\Phi(W_t)=\left\{
	\begin{aligned}
		& 0,  ~~~~~~W_t^{\top}W_t=I, \\
		& +\infty,~\text{otherwise},
	\end{aligned}
	\right.
\end{equation} 
and $ \beta_1, \beta_2, \beta_3 $ are the penalty parameters. 

For ease of description, denote the objective function of problem \eqref{bi-sparse3}  as $\sum_{t=1}^N f(W_t,U_t,V_t,Z) $. Then according to the PAM algorithm \cite{attouch2010proximal,liu2024survey}, it can be solved by iteratively updating $ (W_t,U_t,V_t,Z) $, as shown  in Algorithm \ref{algo}.
\begin{algorithm}[t]
    \caption{Optimization algorithm for FedSSP} \label{algo} 
    \textbf{Input:} $X_t$  \\
    \textbf{Output:} $Z$ \\
    \textbf{For} $k \leq $ maxit \textbf{do}
            \begin{algorithmic}[1]
                \STATE  Update $W_t^{k+1}$ by
                $$\underset{W_t}{\min} ~ f(W_t, U_t^{k},V_t^{k},Z^{k}) + \frac{\tau_{1}}{2}\|W_t-W_t^{k}\|^{2}_{\textrm{F}}$$
                \STATE  Update $U_t^{k+1}$ by 
                $$\underset{U_t}{\min} ~f(W_t^{k+1},U_t, V_t^{k},Z^{k}) + \frac{\tau_{2}}{2}\|U_t-U_t^{k}\|^{2}_{\textrm{F}}$$
                \STATE  Update $V_t^{k+1}$ by 
                $$\underset{V_t}{\min} ~ f(W_t^{k+1},U_t^{k+1}, V_t, Z^{k}) + \frac{\tau_{3}}{2}\|V_t-V_t^{k}\|^{2}_{\textrm{F}}$$
                \STATE  Update $Z^{k+1}$ by 
                $$\underset{Z}{\min} ~ \sum_{t=1}^N f(W_t^{k+1},U_t^{k+1}, V_t^{k+1}, Z) + \frac{\tau_{4}}{2}\|Z-Z^{k}\|^{2}_{\textrm{F}}$$
            \end{algorithmic}
    \textbf{End for}
\end{algorithm}

\subsection{Update $W_t$}
The $W_t$-subproblem can be reduced to 
\begin{equation}\label{solution-w}
 \begin{aligned}
        \underset{W_t\in \textrm{Gr}(d,m)}{\min} ~&  \| (I - W_tW_t^{\top})X_t \|_\textrm{F}^2 + \frac{\beta_1}{2} \| W_t - U_t^k \|_\textrm{F}^2 \\
        &+ \frac{\beta_2}{2} \| W_t - V_t^k \|_\textrm{F}^2 
        + \frac{\beta_3}{2} \| W_t - Z^k \|_\textrm{F}^2\\
        & + \frac{\tau_{1}}{2}\|W_t-W_t^{k}\|^{2}_{\textrm{F}},
    \end{aligned}
\end{equation}
where $\textrm{Gr}(d,m)$ denotes the Grassmann manifold, defined as
\begin{equation}
\textrm{Gr}(d,m) =  \{W_t \in \mathbb{R}^{d \times m} \mid W_t^{\top}W_t = I\}.
\end{equation} 
The objective function can be reformulated by transitioning it from Euclidean space to the Grassmann manifold. Define the objective function in \eqref{solution-w} as
$	g(W_t)$. Thus, the optimization on $\textrm{Gr}(d,m)$ can be equivalent to
\begin{equation}
	\begin{aligned}
		\underset{ W_t \in \textrm{Gr}(d,m)}{\min} ~ & g(W_t).
	\end{aligned}
\end{equation}
 The Euclidean gradient of 
 $g(W_t)$ is
\begin{equation}
    \begin{aligned} 
        \nabla_{W_t} g(W_t) = &- 2S_t W_t + \beta_{1} (W_t-U_t) \\
       & +\beta_{2} (W_t-V_t) + \beta_{3} (W_t-Z),
    \end{aligned}
\end{equation}
where $S_t=X_tX_t^\top.$ Then, the gradient of the objective function, i.e., $\text{grad} g(W_t)$, can be obtained by projecting the Euclidean gradient onto the tangent space of Grassmann manifold at $W_t$, denoted as $\mathcal{P}_{W_t}$. Following \cite{absil2009optimization}, this projection can be expressed as
\begin{equation} \label{grad}
    \begin{aligned}
        \operatorname{grad} g(W_t) & =\mathcal{P}_{W_t} (\nabla g(W_t)) \\
        & =\nabla g(W_t)- W_t \operatorname{sym}(W_t^{\top} \nabla g(W_t)),
    \end{aligned}
\end{equation}
where $\operatorname{sym}(A)=(A+A^{\top}) / 2$ means extracting the symmetric part of $A$. As a result, we do not need to compute the gradient $\operatorname{grad} g(W_t) $. Instead, only the Euclidean gradient $ \nabla g(W_t)$ is needed. Hence, to solve the $W_t$-subproblem, we employ the conjugate gradient method on the Grassmann manifold, as given in Algorithm \ref{GW}.

\begin{algorithm}[t]
\caption{Solution for $W_t$} 
\label{GW}
\textbf{Input:} $X_t, U_t^{k},V_t^{k},Z^{k}$ \\
\textbf{Output:} $W_t$ \\
\textbf{For} $k \leq $  maxit \textbf{do}	
        \begin{algorithmic}[1]
        \STATE Compute the Riemannian gradient $\eta^k$ by \eqref{grad}\\ 
        \STATE Compute the weighted value
        $$\beta^k=\textrm{Tr}(\eta^{k \top} \eta^k) / \textrm{Tr}(\eta^{(k-1)\top} \eta^{k-1}) $$
        \STATE Compute the transport direction
        $$\mathcal{T}_{W^{k-1} \rightarrow W^k}(\xi^{k-1})=\mathcal{P}_{W^k}(\xi^{k-1})$$
        \STATE Compute the search direction
        $$\xi^k=-\eta^k+\beta^k \mathcal{T}_{W^{k-1} \rightarrow W^k}(\xi^{k-1})$$
        \STATE Compute the step size $t^k$ using backtracking line search
        $$f(\mathcal{R}_{W^k}(t^k \xi^k)) \geq f(W^k)+c t^k \operatorname{Tr}(\eta^{k\top} \xi^k)$$
        \STATE Update $W$ using retraction
        $$W^{k+1} = \mathcal{R}_{W^k}(t^k \xi^k)$$
        \end{algorithmic}
\textbf{End for}
\end{algorithm}

\subsection{Update $U_t$}
The $U_t$-subproblem can be simplified to
\begin{equation}\label{solution-u}
	\begin{aligned}
		\underset{U_t}{\min} ~\lambda_2 \|U_t\|_{q}^{q} + \frac{\beta_{1}}{2} \| U_t - W_t^{k+1} \|_\textrm{F}^2 + \frac{\tau_{2}}{2} \| U_t - U_t^{k} \|_\textrm{F}^2.
	\end{aligned}
\end{equation}
Let  $ A_t^{k+1} = \frac{\beta_{1}}{\beta_{1} +\tau_{2}} W_t^{k+1} + \frac{\tau_{2}}{\beta_{1} +\tau_{2}} U_t^{k} $.
Then, problem \eqref{solution-u} becomes
\begin{equation}
    \begin{aligned}
	\underset{U_t}{\min} ~ \lambda_2 \|U_t\|_{q}^{q} + \frac{\beta_{1} + \tau_{2}}{2} \|U_t - A_t^{k+1} \|_\textrm{F}^2.
    \end{aligned}
\end{equation}
Through element-wise decomposition, the optimization problem can be expressed as
\begin{equation}
    \begin{aligned}
     \underset{(u_{ij})_t}{\min}~\lambda_2 |(u_{ij})_t|^{q} + \frac{\beta_{1} + \tau_{2}}{2}((u_{ij})_t - {(a_{ij})}_t^{k+1})^2.
    \end{aligned}
\end{equation}
The following lemma presents the proximal operator solution for  $|\cdot|^q$, and further details are available in \cite{zhou2023revisiting,xiu2025bi}.

\begin{Lemma}\label{lemma}
Consider the proximal operator
\begin{equation}
\operatorname{Prox}(a, \lambda) = \underset{x}{\operatorname{argmin}}\ \lambda |x|^q + \frac{1}{2} (x - a)^2,
\end{equation}
its analytical solution can be expressed as
\begin{equation}
\operatorname{Prox}(a,\lambda) =
\begin{cases}
    \{0\}, & |a| < \kappa(\lambda, q), \\
    \{0, \operatorname{sgn}(a) c(\lambda, q)\}, & |a| = \kappa(\lambda, q), \\
    \{\operatorname{sgn}(a) \varpi_q(|a|)\}, & |a| > \kappa(\lambda, q),
\end{cases}
\end{equation}
where the functions are defined as follows
\begin{equation}
\begin{aligned} 
    c(\lambda, q) &= (2 \lambda (1-q))^{\frac{1}{2-q}},\\
    \kappa(\lambda, q) &= (2-q) \lambda^{\frac{1}{2-q}} (2(1-q))^{\frac{q+1}{q-2}},\\
    \varpi_q(a) &\in \left\{ x \mid x-a + \lambda q \operatorname{sgn}(x) x^{q-1} = 0,\ x > 0 \right\}.
\end{aligned}
\end{equation}
In this formulation 
    	\begin{itemize}    
       	\item  $c(\lambda, q) > 0$ ensures the positivity of the solution; 
       	\item $\kappa(\lambda, q)$ defines the threshold for different regions. 
        \end{itemize}

\end{Lemma}

According to Lemma 1, the solution of problem \eqref{solution-u} can be derived as
\begin{equation}
    \begin{aligned}
        (u_{ij})_t^{k+1} = \operatorname{Prox}((a_{ij})_t^{k+1}, \eta),
    \end{aligned}
\end{equation}
where $\eta = \lambda_2 / (\beta_{1} + \tau_{2})$.

\subsection{Update $V_t$}
The $V_t$-subproblem can be characterized as 
\begin{equation}\label{solution-v}
    \begin{aligned}
       \underset{V_t}{\min}~ \lambda_1 \|V_t\|_{2,p}^{p} + \frac{\beta_{2}}{2} \| V_t - W_t^{k+1} \|_\textrm{F}^2+ \frac{\tau_{3}}{2} \| V_t - V_t^{k} \|_\textrm{F}^2.
    \end{aligned}
\end{equation}
Let $ B_t^{k+1} = \frac{\beta_{2}}{\beta_{2} +\tau_{3}} W_t^{k+1} + \frac{\tau_{3}}{\beta_{2} +\tau_{3}} V_t^k $. Then, it derives
\begin{equation}
    \begin{aligned}
        	\underset{V_t}{\min}~\lambda_1 \|V_t\|_{2,p}^{p} + \frac{\beta_{2} + \tau_{3}}{2} \|V_t - B_t^{k+1} \|_\textrm{F}^2,
    \end{aligned}
\end{equation}
According to the row-wise decomposition, it leads to 
\begin{equation}
    \begin{aligned}
  \underset{(\mathbf{v}^i)_t}{\min}~ \lambda_1 \|(\mathbf{v}^i)_t\|^{p} + \frac{\beta_{2} + \tau_{3}}{2} ((\mathbf{v}^i)_t - (\mathbf{b}^i)_t^{k+1})^{2}.
    \end{aligned}
\end{equation}
According to Lemma \ref{lemma}, the solution of problem \eqref{solution-v} can be also easily obtained as
\begin{equation}
    \begin{aligned}
       (\mathbf{v}^i)_t^{k+1} = \operatorname{Prox}(\|(\mathbf{b}^i)_t^{k+1}\|, \rho)·\frac{(\mathbf{b}^i)_t^{k+1}}{\|(\mathbf{b}^i)_t^{k+1}\|},
    \end{aligned}
\end{equation}
where $\rho= \lambda_1 / (\beta_{2} + \tau_{3})$.

\subsection{Update $Z$}
The $Z$-subproblem can be reformulated as
\begin{equation} 
    \begin{aligned}
        \underset{Z}{\min} ~  \sum_{t=1}^N  \frac{\beta_{3}}{2} \| Z - W_t^{k+1} \|_\textrm{F}^2 + \frac{\tau_{4}}{2} \| Z - Z^{k} \|_\textrm{F}^2.
    \end{aligned}
\end{equation}
A closed-form solution is obtained by taking the derivative as follows
\begin{equation}
    \begin{aligned}
        Z = \frac{\sum_{t=1}^N \beta_{3} W_t^{k+1} + \tau_{4} Z^{k}}{N \beta_{3} + \tau_{4}}.
    \end{aligned}
\end{equation}



	\begin{table}[t] 
		\caption{Distribution of the selected Ton dataset.} \label{dateset}
		\renewcommand\arraystretch{1.2}
		\centering
		\setlength{\tabcolsep}{4mm}{
			\begin{tabular}{|c|c|c|}
				\hline
				\textbf{Class} & \textbf{Training Set} & \textbf{Testing Set} \\ 
				\hline
				\hline
				Normal & 114,956 & 10,000 \\ 
				\hline
				Injection & 12,014 & 7,950 \\ 
				\hline
				Password & 11,956 & 7,891 \\ 
				\hline
				DDoS & 11,902 & 8,084 \\ 
				\hline
				Backdoor & 11,287 & 7,423 \\ 
				\hline
				Scanning & 11,151 & 7,464 \\ 
				\hline
				DoS & 8,781 & 5,888 \\ 
				\hline
				Ransomware & 8,608 & 5,658 \\ 
				\hline
				XSS & 8,706 & 5,813 \\ 
				\hline
				MITM & 653 & 386 \\ 
				\hline
			\end{tabular}
		}
	\end{table}

\section{Case Studies}

In order to demonstrate the effectiveness, this section compares our proposed FedSSP with state-of-the-art methods in an intrusion detection system (IDS) deployment.
The experiments are conducted on a server equipped with an Intel(R) Xeon(R) Platinum 8352V CPU@2.10GHz, Ubuntu 20.04.4 LTS, 64GB RAM, and NVIDIA RTX 4090 GPU.

\subsection{Experimental Settings}

\begin{figure*}[t]  
	\centering
	\begin{minipage}{0.30\textwidth}
		\centering
		\includegraphics[width=\textwidth]{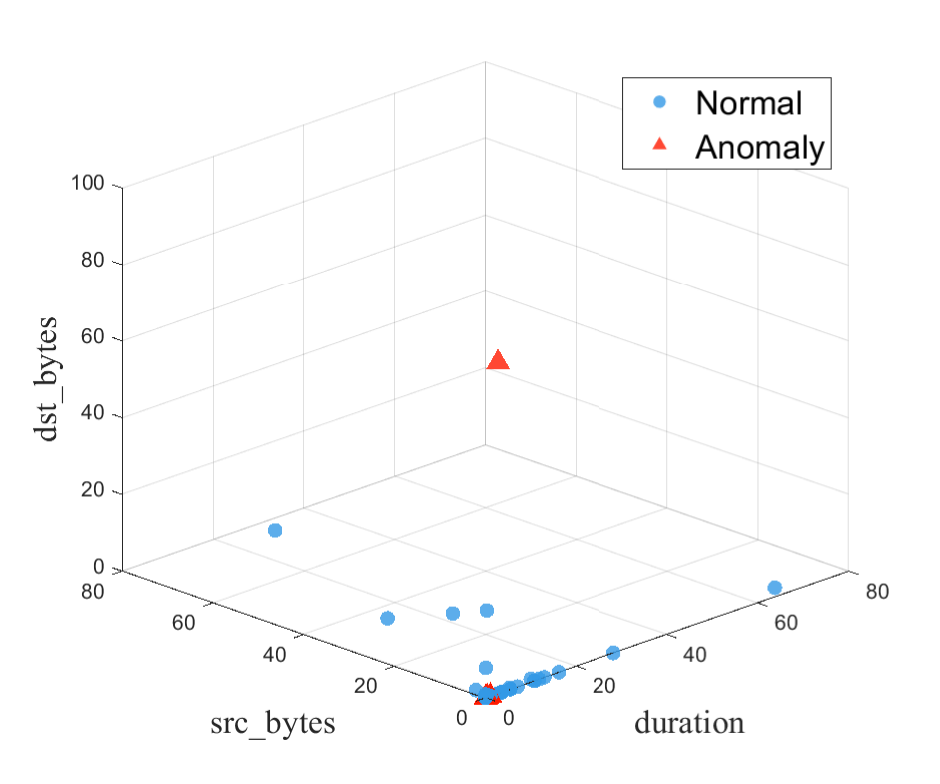}
		\caption{Original}
	\end{minipage}
	\begin{minipage}{0.30\textwidth}
		\centering
		\includegraphics[width=\textwidth]{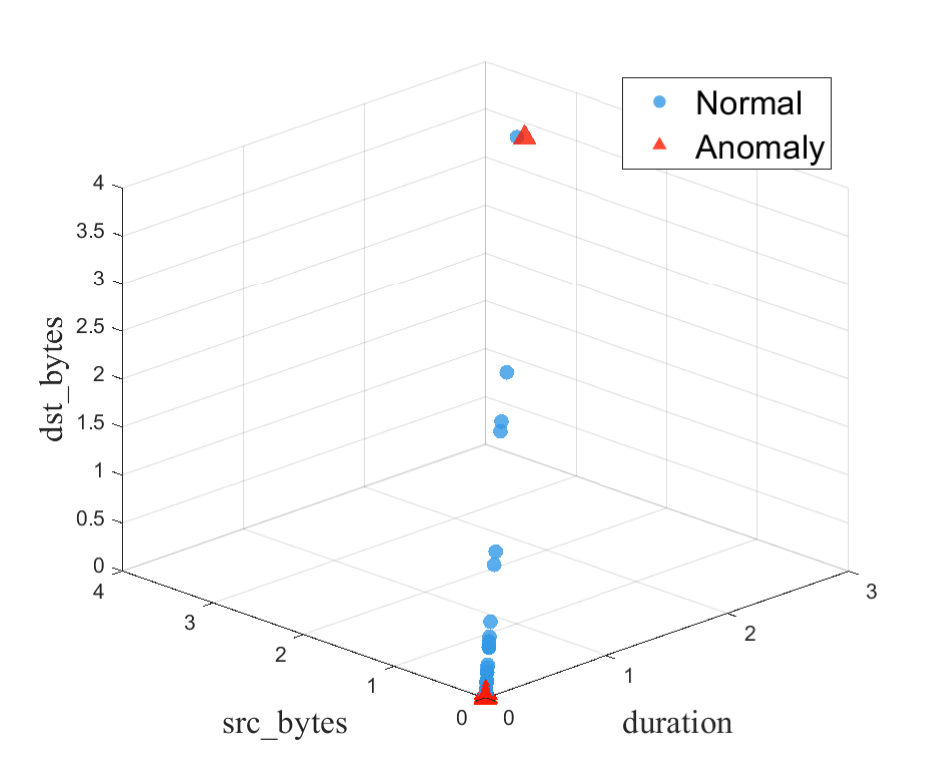}
		\caption{Reconstructed by FedPG}
	\end{minipage}
	\begin{minipage}{0.30\textwidth}
		\centering
		\includegraphics[width=\textwidth]{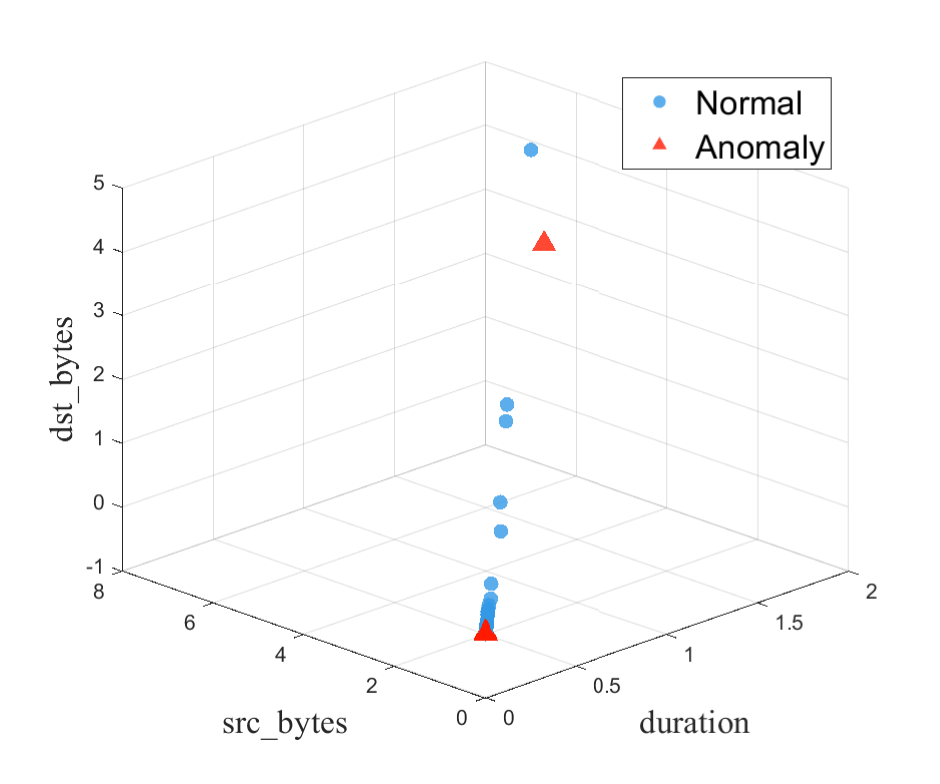}
		\caption{Reconstructed by FedSSP}
	\end{minipage}
	\caption{Visualization of reconstruction results by different methods.} \label{vis}
\end{figure*}

\subsubsection{Dataset Description}

The TON\footnote{https://research.unsw.edu.au/projects/toniot-datasets} dataset is widely used to validate the effectiveness of different algorithms in IoT networks \cite{booij2021ton_iot}. After preprocessing, the training set contains $114,956$ normal samples, the testing set contains $10,000$ normal samples and $56,557$ abnormal samples, all of which have $49$ numerical features. The category distribution of the training set and the testing set is shown in Table \ref{dateset}.



       \begin{table}[t]
       	\centering
       	\caption{Detection performance (\%) on the Ton dataset.}\label{tab2}
       	\renewcommand\arraystretch{1.2}
       	\setlength{\tabcolsep}{4mm}{
       		\begin{tabular}{|c|c|c|c|}
       			\hline
       			\textbf{Metric} & \textbf{FedAE} & \textbf{FedPG} & \textbf{FedSSP}  \\
       			\hline
       			\hline
       			Acc  & 84.97 & 88.61 & 90.10  \\
       			\hline
       			Pre  & 84.97 & 90.56 & 92.08 \\
       			\hline
       			Recall  & 100.00 & 96.67 & 96.67  \\
       			\hline
       			FNR  & 0.00  & 3.33  & 3.33  \\
       			\hline
       			F1   & 91.88 & 93.52 & 94.31  \\
       			\hline
       		\end{tabular}
       	}
       \end{table}
\subsubsection{Evaluation Metrics}
To evaluate the performance of compared methods, the following metrics are considered.
\begin{itemize}
    \item Accuracy (Acc): The proportion of correctly classified records out of the total records.
    \item Precision (Pre): The proportion of correctly identified attacks among all predicted attack records.
    \item Recall (Recall): The proportion of correctly detected attacks among all actual attack records.
    \item False negative rate (FNR): The fraction of actual anomalies incorrectly classified as normal.
    \item F1 Score (F1): The harmonic mean of precision and recall, balancing model performance.
\end{itemize}

\subsubsection{Implementation Issues}
In the IDS deployment, local IoT devices connect to a gateway to collect data for anomaly detection. To reflect the diverse client traffic, the training set is divided into $20$ non-i.i.d. subsets based on ``dst bytes".  The features are normalized by $z$-scores and the hyperparameters are optimized by grid search.

\subsection{Numerical Results}

\subsubsection{Quantitative Analysis}
\label{dp}
This section compares the performance of FedSSP with the following baseline methods: FedPG \cite{nguyen2024federated} and FedAE \cite{ieracitano2020novel}, to test the anomaly detection capability. Specifically, FedPG is a network-free anomaly detection method based on federated PCA, while FedAE is a neural network-based anomaly detection method utilizing autoencoders. 
In FedPG, all records are retrieved to the global server to learn the global variable, regardless of the IDS setting. In FedAE, we only use local records to train the variable for each client and then report the average performance of all clients.

From Table \ref{tab2}, it can be observed that compared with FedPG, although the values of Recall and FNR are the same, ACC is 1.49\% higher, Pre is 1.52\% higher, and F1 is 0.79\% higher. Overall, the performance of the proposed FedSSP is promising, which indicates that the introduced structured sparsity can effectively extract meaningful information from data, thereby improving the performance of anomaly detection in IoT networks.

\begin{figure}[t]
	\centering
	\includegraphics[scale=0.48]{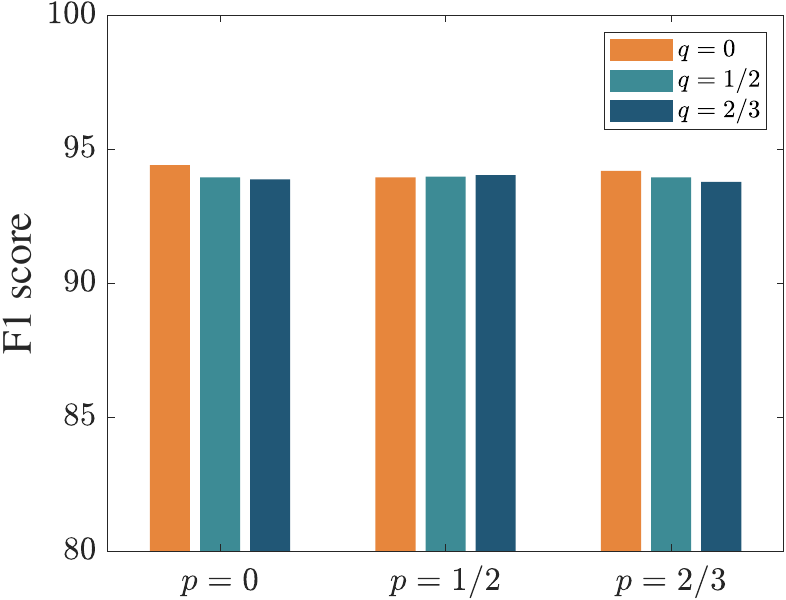} 
	\caption{F1 score (\%) under different $p$ and $q$ values.}

	\label{rounds} \label{pq}
\end{figure}

\subsubsection{Visualization Analysis}

This section visualizes the transformation of DoS traffic records to assess the efficacy of the proposed method.
As shown in Fig. \ref{vis}, three features (duration, src\_bytes, dst\_bytes) are selected and the principal components in FedPG and FedSSP are used to reconstruct the original records. It can be seen that both FedPG and FedSSP can  distinguish between normal and abnormal records, and the reconstruction performance is similar. However,  in local anomaly regions, FedSSP performs better, which is consistent with the improved evaluation metrics discussed before.

\subsubsection{Parameter Analysis}

This section discusses the impact of $p$ and $q$ on the performance of FedSSP. Fig. \ref{pq} illustrates  the F1 scores when $p$ and $q$ are $0, 1/2$, or $2/3$, respectively, where the $x$ axis represents $p$ and bar colors denote $q$. It is found that the F1 score varies between $p$ and $q$ values, and better performance is more likely to be achieved when $q = 0$. The minimum F1 score observed in the bar graph is $93.77\%$, which is higher than FedPG without structured sparsity, i.e., $93.52\%$. This further highlights the effectiveness of our proposed method.

\section{Conclusions}

In this paper, we have proposed a federated structured sparse PCA, which is a novel framework for anomaly detection  in IoT networks. Unlike existing federated PCA, it involves double regularization terms, where the $\ell_{2,p}$-norm can capture row-wise sparsity and the $\ell_{q}$-norm can characterize element-wise sparsity, which complement each other to improve interpretability and detection accuracy. Experimental results show that the proposed method can achieve better performance even compared with the latest FedAE and FedPG.

In the future, we are interested in developing more efficient optimization strategies and integrating the proposed method with contrastive learning for better anomaly detection.

\bibliographystyle{IEEEtran}
\bibliography{mybibfile}

\end{document}